\documentclass[runningheads]{llncs}
\usepackage{graphicx}
\usepackage{verbatim}
\usepackage{hyperref}
\usepackage{float}

\usepackage{hyperref}
\let\svthefootnote\thefootnote
\newcommand\freefootnote[1]{%
  \let\thefootnote\relax%
  \footnotetext{#1}%
  \let\thefootnote\svthefootnote%
}

\begin{document}
\title{An incremental MaxSAT-based model to learn interpretable and 
balanced classification rules}
\titlerunning{An incremental MaxSAT-based model to learn balanced rules}
%
\author{Antônio Carlos Souza Ferreira Júnior \and
Thiago Alves Rocha}
%
%
\institute{Instituto Federal de Educação, Ciência e Tecnologia do Ceará (IFCE), Brazil \\ 
\email{antonio.carlos.souza60@aluno.ifce.edu.br} \\ 
\email{thiago.alves@ifce.edu.br}}
\maketitle              
\begin{abstract}
The increasing advancements in the field of machine learning have led to the 
development of numerous applications that effectively address a wide range of 
problems with accurate predictions. However, in certain cases, accuracy alone may 
not be sufficient. Many real-world problems also demand explanations and 
interpretability behind the predictions. One of the most popular interpretable 
models that are classification rules. This work aims to propose an incremental 
model for learning interpretable and balanced rules based on MaxSAT, called IMLIB. 
This new model was based on two other approaches, one based on SAT and the other 
on MaxSAT. The one based on SAT limits the size of each generated rule, making it 
possible to balance them. We suggest that such a set of rules seem more natural to 
be understood compared to a mixture of large and small rules. The approach based on 
MaxSAT, called IMLI, presents a technique to increase performance that involves 
learning a set of rules by incrementally applying the model in a dataset. Finally, 
IMLIB and IMLI are compared using diverse databases. IMLIB obtained results 
comparable to IMLI in terms of accuracy, generating more balanced rules with 
smaller sizes.

\end{abstract}

\freefootnote{This preprint has not undergone peer review or any post-submission 
improvements or corrections. The Version of Record of this contribution is 
published in Intelligent Systems, LNCS, vol 14195 and is available online at \url{https://doi.org/10.1007/978-3-031-45368-7_15}.}

\section{Introduction}

The success of Machine Learning (ML) in recent years has led to a growing 
advancement in studies in this area 
\cite{carleo2019machine,huang2021power,janiesch2021machine}. Several applications 
have emerged with the aim of circumventing various problems and situations 
\cite{ghassemi2021false,kwekha2021coronavirus,sharma2020machine}. One such problem 
is the lack of explainability of prediction models. This directly affects the 
reliability of using these applications in critical situations involving, for 
example, finance, autonomous systems, damage to equipment, the environment, and 
even lives \cite{biran2017explanation,gunning2019xai,yan2020machine}. That said, 
some works seek to develop approaches that bring explainability to their 
predictions \cite{jimenez2020drug,tjoa2020survey,vilone2021notions}.

Precise predictions with high levels of interpretability are often not a simple 
task. There are some works that try to solve this problem by balancing the accuracy 
of the prediction with the interpretability 
\cite{ghosh2022efficient,ignatiev2021reasoning,mita2020libre,yu2020computing,ghosh2019imli,malioutov2018mlic,lakkaraju2016interpretable}. It can be seen that some of 
these works use approaches based on the Boolean Satisfiability Problem (SAT) and 
the Maximum Boolean Satisfiability Problem (MaxSAT). The choice of these approaches 
to solve this problem has been increasingly recurrent in recent years. The reasons 
can be seen in the results obtained by these models.

SAT-based approaches have been proposed recently 
\cite{rocha2018synthesis,rocha2019synthesis} to learn quantifier-free first-order 
sentences from a set of classified strings. More specifically, given a set of 
classified strings, the goal is to find a first-order sentence over strings of 
minimum size that correctly classifies all the strings. One of the approaches 
demonstrated is SQFSAT (Synthesis of quantifier-free first-order sentences over 
strings with SAT). Upon receiving a set of classified strings, this approach 
generates a quantifier-free first-order sentence over strings in disjunctive normal 
form (DNF) with a given number of terms. What makes this method stand out is the 
fact that we can limit both the number of terms and the number of formulas per term 
in the generated formula. In addition, as the approach generates formulas in DNF, 
each term of the formula can be seen as a rule. Then, for each rule, its 
explanation is the conjunction of formulas in the rule, which can be interesting 
for their interpretability \cite{ignatiev2018sat,rocha2018synthesis}. On the other 
hand, as the model is based on the SAT problem, in certain situations it may bring 
results that are not so interesting in terms of interpretability and efficiency, 
such as in cases where the set of strings is large.

Ghosh, B. et al. created a classification model based on MaxSAT called IMLI 
\cite{ghosh2019imli}. The approach takes a set of classified samples, represented 
by vectors of numerical and categorical data, and generates a set of rules 
expressed in DNF or in conjunctive normal form (CNF) that correctly classifies as 
many samples as possible. In this work, we focus on using IMLI for learning rules 
in DNF. The number of rules in the set of rules can be defined similarly to SQFSAT, 
but IMLI does not consider the number of elements per rule. Although IMLI focuses 
on learning a sparse set of rules, it may obtain a combination of both large and 
small rules. IMLI also takes into account the option of defining a weighting for 
correct classifications. As the weighting increases, the accuracy of the model 
improves, but at the cost of an increase in the size of the generated set of rules. 
The smaller the weighting, the lower the accuracy of the model, but 
correspondingly, the generated set of rules tends to be smaller. Furthermore, IMLI 
uses an incremental approach to achieve better runtime performance. The incremental 
form consists of dividing the set of samples into partitions in order to generate a 
set of rules for each partition from the set of rules obtained in the previous 
partitions.

In this work, we aim to create a new approach for learning interpretable rules 
based on MaxSAT that unites SQFSAT with the incrementability of IMLI. The 
motivation for choosing SQFSAT is the possibility of defining the number of 
literals per clause, allowing us to generate smaller and more balanced rules. The 
choice of IMLI is motivated by its incrementability technique, which allows the 
method to train on large sets of samples efficiently. In addition, we propose a 
technique that reduces the size of the generated rules, removing possible 
redundancies.

This work is divided into 6 sections. In Section~\ref{prelim}, we define the 
general notions and notations. Since all methods presented in this paper use 
Boolean logic, we also define in Section~\ref{prelim} how these methods binarize 
datasets with numerical and categorical data. In 
Section~\ref{rule_learning_methods}, SQFSAT and IMLI are presented, respectively. 
We present SQFSAT in the context of our work where samples consist of binary 
vectors instead of strings, and elements of rules are not first-order sentences 
over strings. In Section~\ref{IMLIB}, our contribution is presented: IMLIB. In 
Section~\ref{exp}, we describe the experiments conducted and the results for the 
comparison of our approach against IMLI. Finally, in the last section, we present 
the conclusions and indicate future work.

\section{Preliminaries}\label{prelim}

We consider the binary classification problem where we are given a set of samples 
and their classifications. The set of samples is represented by a binary matrix of 
size $n \times m$ and their classifications by a vector of size $n$. We call the 
matrix $\textbf{X}$ and the vector $\textbf{y}$. Each row of $\textbf{X}$ is a 
sample of the set and we will call it $\textbf{X}_i$ with $i \in \{1,...,n\}$. To 
represent a specific value of $\textbf{X}_i$, we will use $x_{i,j}$ with $j \in 
\{1,...,m\}$. Each column of $\textbf{X}$ has a label representing a feature and 
the label is symbolized by $x^j$. To represent a specific value of $\textbf{y}$, 
we will use $y_i$.

To represent the opposite value of $y_i$, that is, if it is $1$ the opposite value 
is $0$ and vice versa, we use $\lnot y_i$. Therefore, we will use the symbol $\lnot 
\textbf{y}$ to represent $\textbf{y}$ with all opposite values. To represent the 
opposite value of $x_{i,j}$, we use $\lnot x_{i,j}$. Therefore, we will use the 
symbol $\lnot \textbf{X}_i$ to represent $\textbf{X}_i$ with all opposite values.
Each label also has its opposite label which is symbolized by $\lnot x^j$.

A partition of $\textbf{X}$ is represented by $\textbf{X}^t$ with $t \in 
\{1,...,p\}$, where $p$ is the number of partitions. Therefore, the partitions of 
vector $\textbf{y}$ are represented by $\textbf{y}^t$. Each element of $\textbf{y}$ 
is symbolized by $y_i$ and represents the class value of sample $\textbf{X}_i$. We 
use $\mathcal{E}^- = \{X_i \mid y_i = 0, 1 \leq i \leq n\}$ and $\mathcal{E}^+ = 
\{X_i \mid y_i = 1, 1 \leq i \leq n\}$. To represent the size of these sets, that 
is, the number of samples contained in them, we use the notations: 
$|\mathcal{E}^-|$ and $|\mathcal{E}^+|$.

\begin{example}\label{Example: Set of samples}
Let $\textbf{X}$ be the set of samples 

$$\textbf{X} = 
\left[\begin{array}{ccc}
0&0&1\\
0&1&0\\
0&1&1\\
1&0&0
\end{array}\right]$$

\noindent and their classifications $\textbf{y} = [1, 0, 0, 1]$. The samples 
$\textbf{X}_i$ are: $\textbf{X}_1 = [0, 0, 1], ..., \textbf{X}_4 = [1, 0, 0]$. 
The values of each sample $x_{i,j}$ are: $x_{1,1}=0, x_{1,2}=0, x_{1,3}=1, 
x_{2,1}=0, ..., x_{4,3}=0$. The class values $y_i$ of each sample are: 
$y_1=1, ..., y_4=1$. We can divide $\textbf{X}$ into two partitions in several 
different ways, one of which is: 
$\textbf{X}^1 = 
\left[\begin{array}{ccc}
0&1&1\\
0&1&0
\end{array}\right]$, $\textbf{y}^1 = [0, 0]$,
$\textbf{X}^2 = 
\left[\begin{array}{ccc}
1&0&0\\
0&0&1
\end{array}\right]$ e $\textbf{y}^2 = [1, 1]$.
\end{example}

\begin{example}\label{Example: Set of samples opposite}
Let $\textbf{X}$ be the set of samples from Example~\ref{Example: Set of samples}, 
then 

$$\lnot \textbf{X} = 
\left[\begin{array}{ccc}
1&1&0\\
1&0&1\\
1&0&0\\
0&1&1
\end{array}\right]$$

\noindent and  $\lnot \textbf{y} = [0, 1, 1, 0]$. The samples $\lnot \textbf{X}_i$ 
are: $\lnot \textbf{X}_1 = [1, 1, 0], ..., \lnot \textbf{X}_4 = [0, 1, 1]$. The 
values of each sample $\lnot x_{i,j}$ are: $\lnot x_{1,1}=1, \lnot x_{1,2}=1, 
\lnot x_{1,3}=0, \lnot x_{2,1}=1, ..., \lnot x_{4,3}=1$. The class values of each 
sample $\lnot y_i$ are: $\lnot y_1=0, ..., \lnot y_4=0$. We can divide $\lnot 
\textbf{X}$ in partitions as in Example~\ref{Example: Set of samples}: 
$\lnot \textbf{X}^1 = 
\left[\begin{array}{ccc}
1&0&0\\
1&0&1
\end{array}\right]$, $\lnot \textbf{y}^1 = [1, 1]$,
$\lnot \textbf{X}^2 = 
\left[\begin{array}{ccc}
0&1&1\\
1&1&0
\end{array}\right]$ e $\lnot \textbf{y}^2 = [0, 0]$.
\end{example}

\noindent We define a set of rules being the disjunction of rules and is 
represented by $\textbf{R}$. A rule is a conjunction of one or more features. Each 
rule in $\textbf{R}$ is represented by $R_o$ with $o \in \{1,...,k\}$, where $k$ is 
the number of rules. Moreover, $\textbf{R}(\textbf{X}_i)$ represents the 
application of $\textbf{R}$ to $\textbf{X}_i$. The notations $|\textbf{R}|$ and 
$|R_o|$ are used to represent the number of features in $\textbf{R}$ and $R_o$, 
respectively.

\begin{example}\label{Example: Set of rules}
Let $x^1=$ \textit{Man}, $x^2=$ \textit{Smoke}, $x^3=$ \textit{Hike} be labels of 
features. Let $\textbf{R}$ be the set of rules $\textbf{R} =$ (\textit{Man}) $\lor$ 
(\textit{Smoke} $\land$ $\lnot$\textit{Hike}). The rules $R_o$ are: $R_1=$ 
(\textit{Man}) and $R_2=$ (\textit{Smoke} $\land$ $\lnot$\textit{Hike}). The 
application of $\textbf{R}$ to $\textbf{X}_i$ is represented as follows: $\textbf{R}
(\textbf{X}_i) = x_{i,1} \lor (x_{i,2} \land \lnot x_{i,3})$. For example, Let 
$\textbf{X}$ be the set of samples from Example~\ref{Example: Set of samples}, 
then: $\textbf{R}(\textbf{X}_1) = x_{1,1} \lor (x_{1,2} \land \lnot x_{1,3}) = 0 
\lor (0 \land 0) = 0$. Moreover, we have that $|\textbf{R}| = 3$, $|R_1| = 1$ and 
$|R_2| = 2$.
\end{example}

As we assume a set of binary samples, we need to perform some preprocessing. 
Preprocessing consists of binarizing a set of samples with numerical or categorical 
values. The algorithm divides the features into four types: constant, where all 
samples have the same value; binary, where there are only two distinct variations 
among all the samples for the same value; categorical, when the feature does not 
fit in constant and binary and its values are three or more categories; ordinal, 
when the feature does not fit into constant and binary and has numerical values.

When the feature type is constant, the algorithm discards that feature. This 
happens due to the fact that a feature common to all samples makes no difference in 
the generated rules. When the type is binary, one of the feature variations will 
receive $0$ and the other $1$ as new values. If the type is categorical, we employ 
the widely recognized technique of one-hot encoding. Finally, for the ordinal type 
feature, a quantization is performed, that is, the variations of this feature are 
divided into quantiles. With this, Boolean values are assigned to each quantile 
according to the original value.

We use SAT and MaxSAT solvers to implement the methods presented in this work. A 
solver receives a formula in CNF, for example: $(p \lor q) \land (q \lor \lnot p)$. 
Furthermore, a MaxSAT solver receives weights that will be assigned to each clause 
in the formula. A clause is the disjunction of one or more literals. The weights 
are represented by $W(Cl) = w$ where $Cl$ is one or more clauses and $w$ represents 
the weight assigned to each one of them. A SAT solver tries to assign values to the 
literals in such a way that all clauses are satisfied. A MaxSAT solver tries to 
assign values to the literals in a way that the sum of the weights of satisfied 
clauses is maximum. Clauses with numerical weights are considered \textit{soft}. 
The greater the weight, the greater the priority of the clause to be satisfied. 
Clauses assigned a weight of $\infty$ are considered \textit{hard} and must be 
satisfied.

\section{Rule learning with SAT and MaxSAT}\label{rule_learning_methods}

\subsection{SQFSAT}\label{l-QDNFSAT}

SQFSAT is a SAT-based approach that, given \textbf{X}, \textbf{y}, $k$ and the 
number of features per rule $l$, tries to find a set of rules $\textbf{R}$ with $k$ 
rules and at most $l$ features per rule that correctly classify all samples 
$\textbf{X}_i$, that is, $\textbf{R}(\textbf{X}_i) = y_i$ for all $i$. In general, 
the approach takes its parameters $\textbf{X}$, $\textbf{y}$, $k$ and $l$ and 
constructs a CNF formula to apply it to a SAT solver, which returns an answer that 
is used to get $\textbf{R}$.

The construction of the SAT clauses is defined by propositional variables: 
$u_{o,d}^j$, $p_{o,d}$, $u_{o,d}^*$, $e_{o,d,i}$ and $z_{o,i}$, for $d \in 
\{1,...,l\}$. If the valuation of $u_{o,d}^j$ is true, it means that $j$th feature 
label will be the $d$th feature of the rule $R_o$. Furthermore, if $p_{o,d}$ is 
true, it means that the $d$th feature of the rule $R_o$ will be $x^j$, in other 
words, will be positive. Otherwise, it will be negative: $\lnot x^j$. If 
$u_{o,d}^*$ is true, it means that the $d$th feature is skipped in the rule $R_o$. 
In this case, we ignore $p_{o,d}$. If $e_{o,d,i}$ is true, then the $d$th feature 
of rule $R_o$ contributes to the correct classification of the $i$th sample. If 
$z_{o,i}$ is true, then the rule $R_o$ contributes to the correct classification of 
the $i$th sample. That said, below, we will show the constraints formulated in the 
model for constructing the SAT clauses.

Conjunction of clauses that guarantees that exactly one $u_{o,d}^j$ is true for the 
$d$th feature of the rule $R_o$:
\begin{equation} \label{Constraint 1: l-QDNFSAT}
A = \bigwedge_{o \in \{1,...,k\} \atop d \in \{1,...,l\}}
\bigvee_{j \in \{1,...,m,*\}} u_{o,d}^j
\end{equation}

\begin{equation} \label{Constraint 2: l-QDNFSAT}
B = \bigwedge_{o \in \{1,...,k\} \atop{d \in \{1,...,l\} 
\atop j,j' \in \{1,...,m,*\}, j \neq j'}} (\lnot u_{o,d}^j \lor 
\lnot u_{o,d}^{j'})
\end{equation}

Conjunction of clauses that ensures that each rule has at least one feature:
\begin{equation} \label{Constraint 3: l-QDNFSAT}
C = \bigwedge_{o \in \{1,...,k\}} \bigvee_{d \in \{1,...,l\}} 
\lnot u_{o,d}^*
\end{equation}

We will use the symbol $s_{o,d,i}^j$ to represent the value of the $i$th sample in 
the $j$th feature label of $\textbf{X}$. If this value is $1$, it means that if the 
$j$th feature label is in the $d$th position of the rule $R_o$, then it contributes 
to the correct classification of the $i$th sample. Therefore, $s_{o,d,i}^j = 
e_{o,d,i}$. Otherwise, $s_{o,d,i}^j = \lnot e_{o,d,i}$. That said, the following 
conjunction of formulas guarantees that $e_{o,d,i}$ is true if the $j$th feature in 
the $o$th rule contributes to the correct classification of the sample 
$\textbf{X}_i$:
\begin{equation} \label{Constraint 4: l-QDNFSAT}
D = \bigwedge_{o \in \{1,...,k\} \atop{d \in \{1,...,l\} 
\atop{j \in \{1,...,m\} \atop i \in \{1,...,n\}}}} u_{o,d}^j 
\rightarrow (p_{o,d} \leftrightarrow s_{o,d,i}^j)
\end{equation}

Conjunction of formulas guaranteeing that if the $d$th feature of a rule is 
skipped, then the classification of this rule is not interfered by this feature:
\begin{equation} \label{Constraint 5: l-QDNFSAT}
E = \bigwedge_{o \in \{1,...,k\} \atop{d \in \{1,...,l\} 
\atop i \in \{1,...,n\}}} u_{o,d}^* \rightarrow e_{o,d,i}
\end{equation}

Conjunction of formulas indicating that $z_{o,i}$ will be set to true if all the 
features of rule $R_o$ contribute to the correct classification of sample 
$\textbf{X}_i$:
\begin{equation} \label{Constraint 6: l-QDNFSAT}
F = \bigwedge_{o \in \{1,...,k\}} \bigwedge_{i \in \{1,...,n\}} 
z_{o,i} \leftrightarrow \bigwedge_{d \in \{1,...,l\}} e_{o,d,i}
\end{equation}

Conjunction of clauses that guarantees that $\textbf{R}$ will correctly classify 
all samples:
\begin{equation} \label{Constraint 7: l-QDNFSAT}
G = \bigwedge_{i \in \mathcal{E}^+} \bigvee_{o \in \{1,...,k\}} 
z_{o,i}
\end{equation}

\begin{equation} \label{Constraint 8: l-QDNFSAT}
H = \bigwedge_{i \in \mathcal{E}^-} \bigwedge_{o \in \{1,...,k\}} 
\lnot z_{o,i}
\end{equation}

Next, the formula $Q$ below is converted to CNF. Then, finally, we have the SAT 
query that is sent to the solver.
\begin{equation} \label{Query SAT: l-QDNFSAT}
Q = A \land B \land C \land D \land E \land F \land G \land H
\end{equation}

\subsection{IMLI}\label{IMLI}

IMLI is an incremental approach based on MaxSAT for learning interpretable rules. 
Given $\textbf{X}$, $\textbf{y}$, $k$, and a weight $\lambda$, the model aims to 
obtain the smallest set of rules $\textbf{M}$ in CNF that correctly classifies as 
many samples as possible, penalizing classification errors with $\lambda$. In 
general, the method solves the optimization problem $\min_{\textbf{M}} 
\{|\textbf{M}| + \lambda|\mathcal{E}_M| \mid \mathcal{E}_M = \{\textbf{X}_i \mid 
\textbf{M}(\textbf{X}_i) \neq y_i\}\}$, where $|\textbf{M}|$ represents the number 
of features in $\textbf{M}$ and $\textbf{M}(\textbf{X}_i)$ denotes the application 
of the set of rules $\textbf{M}$ to $\textbf{X}_i$. Therefore, the approach takes 
its parameters $\textbf{X}$, $\textbf{y}$, $k$ and $\lambda$ and constructs a 
MaxSAT query to apply it to a MaxSAT solver, which returns an answer that is used 
to generate $\textbf{M}$. Note that IMLI generates set of rules in CNF, whereas our 
objective is to obtain sets of rules in DNF. For that, we will have to use as 
parameter $\lnot \textbf{y}$ instead of $\textbf{y}$ and negate the set of rules 
$\textbf{M}$ to obtain a set of rules $\textbf{R}$ in DNF.

The construction of the MaxSAT clauses is defined by propositional variables: 
$b_o^v$ and $\eta_i$, for $v \in \{1,...,2m\}$. The $v$ ranges from $1$ to $2m$, as 
it also considers opposite features. If the valuation of $b_o^v$ is true and $v 
\leq m$, it means that feature $x^v$ will be in the rule $M_o$, where $M_o$ is the 
$o$th rule of $\textbf{M}$. If the valuation of the $b_o^v$ is true and $v > m$, it 
means that feature $\lnot x^{v - m}$ will be in the rule $M_o$. If the valuation of 
$\eta_i$ is true, it means that sample $\textbf{X}_i$ is not classified correctly, 
that is, $\textbf{M}(\textbf{X}_i) \neq y_i$. That said, below, we will show the 
constraints for constructing MaxSAT clauses.

Constraints that represent that the cost of a misclassification is $\lambda$:
\begin{equation} \label{Constraint 1: IMLI}
A = \bigwedge_{i \in \{1,...,n\}} \lnot \eta_i, W(A) = \lambda
\end{equation}

Constraints that represent that the model tries to insert as few features as 
possible in $\textbf{M}$, taking into account the weights of all clauses:
\begin{equation} \label{Constraint 2: IMLI}
B = \bigwedge_{v \in \{1,...,2m\} \atop o \in \{1,...,k\}} 
\lnot b_o^v, W(B) = 1
\end{equation}

Even though the constraints in~\ref{Constraint 2: IMLI} prioritize learning sparse 
rules, they do so by directing attention to the overall set of rules, i.e. in the 
total number of features in $\textbf{M}$. Then, IMLI may generate a set of rules 
that comprises a combination of both large and small rules. In our approach 
presented in Section~\ref{IMLIB}, we address this drawback by limiting the number 
of features in each rule.

We will use $\textbf{L}_o$ to represent the set of variables $b_o^v$ of a rule 
$M_o$, that is, $\textbf{L}_o = \{b_o^v\, | v \in \{1, ...,2m\}\}$, for $o \in 
\{1,...,k\}$. To represent the concatenation of two samples, we will use the symbol 
$\cup$. We also use the symbol $@$ to represent an operation between two vectors of 
the same size. The operation consists of applying a conjunction between the 
corresponding elements of the vectors. Subsequently, a disjunction between the 
elements of the result is applied. The following example illustrates how these 
definitions will be used:

\begin{example}\label{Example: @ operation}
Let be $\textbf{X}_4$ as in Example \ref{Example: Set of samples}, $\textbf{X}_4 
\cup \lnot \textbf{X}_4 = [1,0,0,0,1,1]$ and $\textbf{L}_o = 
[b_o^1,b_o^2,b_o^3,b_o^4,b_o^5,b_o^6]$. Therefore, $(\textbf{X}_4 \cup \lnot 
\textbf{X}_4) @ \textbf{L}_o = (x_{4,1} \land b_o^1) \lor (x_{4,2} \land b_o^2) 
\lor ... \lor (\lnot x_{4,6} \land b_o^6) = (1 \land b_o^1) \lor (0 \land b_o^2) 
\lor ... \lor (1 \land b_o^6) = b_o^1 \lor b_o^5 \lor b_o^6$.
\end{example}

The objective of this operation is to generate a disjunction of variables that 
indicates if any of the features associated with these variables are present in 
$M_o$, then sample $\textbf{X}_i$ will be correctly classified by $M_o$. Now, we 
can show the formula that guarantees that if $\eta_i$ is false, then $\textbf{M}
(\textbf{X}_i) = y_i$:
\begin{equation} \label{Constraint 3: IMLI}
C = \bigwedge_{\ i\ \in\ \{1, ..., n\}} \lnot\eta_i \rightarrow (y_i 
\leftrightarrow \bigwedge_{o \in \{1,...,k\}} ((\textbf{X}_i \cup 
\lnot \textbf{X}_i) @ \textbf{L}_o)), W(C) = \infty
\end{equation}

We can see that $C$ is not in CNF. Therefore, formula $Q$ below must be converted 
to CNF. With that, finally, we have the MaxSAT query that is sent to the solver.
\begin{equation} \label{Query MaxSAT: IMLI}
Q = A \land B \land C
\end{equation}

\noindent The set of samples $\textbf{X}$, in IMLI, can be divided into $p$ 
partitions: $\textbf{X}^1$, $\textbf{X}^2$, ..., $\textbf{X}^p$. Each partition, 
but the last one, contains the same values of $|\mathcal{E}^-|$ and 
$|\mathcal{E}^+|$. Also, the samples are randomly distributed across the 
partitions. Partitioning aims to make the model perform better in generating the 
set of rules $\textbf{M}$. Thus, the conjunction of clauses will be created from 
each partition $\textbf{X}^t$ in an incremental way, that is, the set of rules 
$\textbf{M}$ obtained by the current partition will be reused for the next 
partition. In the first partition, constraints (\ref{Constraint 1: IMLI}), 
(\ref{Constraint 2: IMLI}), (\ref{Constraint 3: IMLI}) are created in exactly the 
same way as described. From the second onwards, (\ref{Constraint 2: IMLI}) is 
replaced by the following constraints:
\begin{equation} \label{Constraint 2.1: IMLI}
B' = \bigwedge_{v \in \{1,...,2m\} \atop o \in \{1,...,k\}} 
\left\{\begin{array}{ll}
b_o^v \textrm{, if $b_o^v$ is true in the previous partition};\\
\lnot b_o^v \textrm{, otherwise}.
\end{array}
\right., W(B') = 1
\end{equation}

The IMLI also has a technique for reducing the size of the generated set of rules. 
The technique removes possible redundancies in ordinal features as the one in 
Example \ref{Example: Feature redundancy in the same rule}. In the original 
implementation of the model, the technique is applied at the end of each partition. 
In our implementation for the experiments in Section~\ref{exp}, this technique 
is applied only at the end of the last partition. The reason for this is training 
performance.

\begin{example}\label{Example: Feature redundancy in the same rule}
Let $\textbf{R}$ be the following set of rules with redundancy in the same 
rule: 
$$(\textit{Age}>18 \land \textit{Age}>20) \lor (\textit{Height}\leq2).$$

\noindent Then, the technique removes the redundancy and the following set 
of rules is obtained:
$$(\textit{Age}>20) \lor (\textit{Height}\leq2).$$
\end{example}

\section{IMLIB}\label{IMLIB}

In this section, we will present our method IMLIB which is an incremental version 
of SQFSAT based on MaxSAT. IMLIB also has a technique for reducing the size of the 
generated set of rules. Therefore, our approach partitions the set of samples 
$\textbf{X}$. Moreover, our method has one more constraint and weight on all 
clauses. With that, our approach receives five input parameters $\textbf{X}$, 
$\textbf{y}$, $k$, $l$, $\lambda$ and tries to obtain the smallest $\textbf{R}$ 
that correctly classifies as many samples as possible, penalizing classification 
errors with $\lambda$, that is, $\min_{\textbf{R}} \{|\textbf{R}| + 
\lambda|\mathcal{E}_R| \mid \mathcal{E}_R = \{\textbf{X}_i \mid \textbf{R}
(\textbf{X}_i) \neq y_i\}\}$. That said, below, we will show the constraints of our 
approach for constructing MaxSAT clauses.

Constraints that guarantee that exactly only one $u_{o,d}^j$ is true for the $d$th 
feature of the rule $R_o$:
\begin{equation} \label{Constraint 1: IMLIB}
A = \bigwedge_{o \in \{1,...,k\} \atop d \in \{1,...,l\}}
\bigvee_{j \in \{1,...,m,*\}} u_{o,d}^j, W(A) = \infty
\end{equation}

\begin{equation} \label{Constraint 2: IMLIB}
B = \bigwedge_{o \in \{1,...,k\} \atop{d \in \{1,...,l\} 
\atop j,j' \in \{1,...,m,*\}, j \neq j'}} \lnot u_{o,d}^j \lor 
\lnot u_{o,d}^{j'}, W(B) = \infty
\end{equation}

Constraints representing that the model will try to insert as few features as 
possible in $\textbf{R}$:
\begin{equation} \label{Constraint 3: IMLIB}
C = \bigwedge_{o \in \{1,...,k\} \atop{d \in \{1,...,l\} \atop 
j \in \{1,...,m\}}} \lnot u_{o,d}^j \land \bigwedge_{o \in \{1,...,k\} \atop{d \in 
\{1,...,l\}}} u_{o,d}^*, W(C) = 1
\end{equation}

Conjunction of clauses that guarantees that each rule has at least one feature:
\begin{equation} \label{Constraint 4: IMLIB}
D = \bigwedge_{o \in \{1,...,k\}} \bigvee_{d \in \{1,...,l\}} 
\lnot u_{o,d}^*, W(D) = \infty
\end{equation}

The following conjunction of formulas ensures that $e_{o,d,i}$ is true if the $j$th 
feature label in the $o$th rule contributes to correctly classify sample 
$\textbf{X}_i$:
\begin{equation} \label{Constraint 5: IMLIB}
E = \bigwedge_{o \in \{1,...,k\} \atop{d \in \{1,...,l\} 
\atop{j \in \{1,...,m\} \atop i \in \{1,...,n\}}}} u_{o,d}^j 
\rightarrow (p_{o,d} \leftrightarrow s_{o,d,i}^j), W(E) = \infty
\end{equation}

Conjunction of formulas guaranteeing that the classification of a specific rule 
will not be interfered by skipped features in the rule:
\begin{equation} \label{Constraint 6: IMLIB}
F = \bigwedge_{o \in \{1,...,k\} \atop{d \in \{1,...,l\} 
\atop i \in \{1,...,n\}}} u_{o,d}^* \rightarrow e_{o,d,i}, W(F) = 
\infty
\end{equation}

Conjunction of formulas indicating that the model assigns true to $z_{o,i}$ if all 
the features of rule $R_o$ support the correct classification of sample 
$\textbf{X}_i$:
\begin{equation} \label{Constraint 7: IMLIB}
G = \bigwedge_{o \in \{1,...,k\}} \bigwedge_{i \in \{1,...,n\}} 
z_{o,i} \leftrightarrow \bigwedge_{d \in \{1,...,l\}} e_{o,d,i}, 
W(G) = \infty
\end{equation}

Conjunction of clauses designed to generate a set of rules $\textbf{R}$ that 
correctly classify as many samples as possible:
\begin{equation} \label{Constraint 8: IMLIB}
H = \bigwedge_{i \in \mathcal{E}^+} \bigvee_{o \in \{1,...,k\}} 
z_{o,i}, W(H) = \lambda
\end{equation}

\begin{equation} \label{Constraint 9: IMLIB}
I = \bigwedge_{i \in \mathcal{E}^-} \bigwedge_{o \in \{1,...,k\}} 
\lnot z_{o,i}, W(I) = \lambda
\end{equation}

Finally, after converting formula $Q$ below to CNF, we have the MaxSAT query that 
is sent to the solver.
\begin{equation} \label{Query MaxSAT: IMLIB}
Q = A \land B \land C \land D \land E \land F \land G \land H 
\land I
\end{equation}

\noindent IMLIB can also partition the set of samples $\textbf{X}$ in the same way 
IMLI. Therefore, all constraints described above are applied in the first 
partition. Starting from the second partition, the constraints in 
(\ref{Constraint 3: IMLIB}) are replaced by the following constraints:
\begin{equation} \label{Constraint 3.1: IMLIB}
C' = \bigwedge_{o \in \{1,...,k\} \atop{d \in \{1,...,l\} \atop 
j \in \{1,...,m,*\}}} 
\left\{\begin{array}{ll}
u_{o,d}^j \textrm{, if $u_{o,d}^j$ is true in the previous}\\
{partition};\\ \lnot u_{o,d}^j \textrm{, otherwise}.
\end{array}
\right., W(C') = 1
\end{equation}

\noindent IMLIB also has a technique for reducing the size of the generated set of 
rules demonstrated in Example \ref{Example: Feature redundancy in the same rule}. 
Moreover, we added two more cases which are described in Example 
\ref{Example: Opposite features in the same rule} and Example 
\ref{Example: The same feature in the same rule}.

\begin{example}\label{Example: Opposite features in the same rule}
Let $\textbf{R}$ be the following set of rules with opposite features in the same 
rule: $$(\textit{Age}>20) \lor (\textit{Height}\leq2 \land \textit{Height}>2) \lor 
(\textit{Hike} \land \textit{Not Hike}).$$

\noindent Therefore, the technique removes rules with opposite features in the same 
rule obtaining the following set of rules: $$(\textit{Age}>20).$$
\end{example}

\begin{example}\label{Example: The same feature in the same rule}
Let $\textbf{R}$ be the following set of rules with the same feature occurring 
twice in a rule: $$(\textit{Hike} \land \textit{Hike}) \lor (\textit{Age}>20).$$

\noindent Accordingly, our technique for removing redundancies eliminates 
repetitive features, resulting in the following set of rules: $$(\textit{Hike}) 
\lor (\textit{Age}>20).$$
\end{example}

\section{Experiments}\label{exp}

\begin{table}[t]
    \centering
    \caption{
        Databases information.
    }\label{Table: Databases information}
    \begin{tabular}{|c|c|c|c|c|}
        \hline
            Databases & Samples & $|\mathcal{E}^-|$ & $|\mathcal{E}^+|$ & 
            Features \\
        \hline
        \hline
            lung cancer & 59 & 31 & 28 & 6\\
        \hline
            iris & 150 & 100 & 50 & 4\\
        \hline
            parkinsons & 195 & 48 & 147 & 22 \\
        \hline
            ionosphere & 351 & 126 & 225 & 33 \\
        \hline
            wdbc & 569 & 357 & 212 & 30 \\
        \hline
            transfusion & 748 & 570 & 178 & 4\\
        \hline
            pima & 768 & 500 & 268 & 8 \\
        \hline
            titanic & 1309 & 809 & 500 & 6 \\
        \hline
            depressed & 1429 & 1191 & 238 & 22 \\
        \hline
            mushroom & 8112 & 3916 & 4208 & 22 \\
        \hline
    \end{tabular}
\end{table}

In this section, we present the experiments we conducted to compare our method 
IMLIB against IMLI. The two models were implemented\footnote{Source code of IMLIB 
and the implementation of the tests performed can be found at the link: 
\url{https://github.com/cacajr/decision_set_models}} with Python and MaxSAT solver 
RC2 \cite{ignatiev2019rc2}. The experiments were carried out on a machine with the 
following configurations: Intel(R) Core(TM) i5-4460 3.20GHz processor, and 12GB of 
RAM memory. Ten databases from the UCI repository \cite{Dua:2019} were used to 
compare IMLI with IMLIB. Information on the datasets can be seen in Table 
\ref{Table: Databases information}. Databases that have more than two classes were 
adapted, considering that both models are designed for binary classification. For 
purposes of comparison, we measure the following metrics: number of rules, size of 
the set of rules, size of the largest rule, accuracy on test data and training 
time. The number of rules, size of the set of rules, and size of the largest rule 
can be used as interpretability metrics. For example, a set of rules with few rules 
and small rules is more interpretable than one with many large rules.

Each dataset was split into $80\%$ for training and $20\%$ for testing. Both models 
were trained and evaluated using the same training and test sets, as well as the 
same random distribution. Then, the way the experiments were conducted ensured that 
both models had exactly the same set of samples to learn the set of rules. 

For both IMLI and IMLIB, we consider parameter configurations obtained by combining 
values of: $k \in \{1,2,3\}$, $\lambda \in \{5,10\}$ and $lp \in \{8,16\}$, where 
$lp$ is the number of samples per partition. Since IMLIB has the maximum number of 
features per rule $l$ as an extra parameter, for each parameter configuration of 
IMLI and its corresponding $\mathbf{R}$, we considered $l$ ranging from $1$ to one 
less than the size of the largest rule in $\mathbf{R}$. Thus, the value of $l$ that 
resulted in the best test accuracy was chosen to be compared with IMLI. Our 
objective is to evaluate whether IMLIB can achieve higher test accuracy compared to 
IMLI by employing smaller and more balanced rules. Furthermore, it should be noted 
that this does not exclude the possibility of our method generating sets of rules 
with larger sizes than IMLI. 

For each dataset and each parameter configuration of $k$, $\lambda$ and $lp$, we 
conducted ten independent realizations of this experiment. For each dataset, the 
parameter configuration with the best average of test accuracy for IMLI was chosen 
to be inserted in Table \ref{Table: All results IMLI}. For each dataset, the 
parameter configuration with the best average of test accuracy for IMLIB was chosen 
to be inserted in Table \ref{Table: All results IMLIB}. The results presented in 
both tables are the average over the ten realizations.

\begin{table}[t]
    \centering
    \scriptsize
    \caption{
        Comparison between IMLI and IMLIB in different databases with the 
        IMLI configuration that obtained the best result in terms of 
        accuracy. The column Training time represents the training time in 
        seconds.
    }\label{Table: All results IMLI}
    \begin{tabular}{|c|c|c|c|c|c|c|}
    \hline
        Databases & Models & \begin{tabular}{c} Number of\\ rules \end{tabular} 
        & $|\textbf{R}|$ & \begin{tabular}{c} Largest rule\\ size \end{tabular} 
        & Accuracy & Training time \\ \hline\hline
        lung cancer 
          & IMLI & \textbf{2.00 $\pm$ 0.00} & 3.60 $\pm$ 0.84 & 2.20 $\pm$ 0.63 & \textbf{0.93 $\pm$ 0.07} & \textbf{0.0062 $\pm$ 0.0016} \\
        ~ & IMLIB & \textbf{2.00 $\pm$ 0.00} & \textbf{2.20 $\pm$ 0.63} & \textbf{1.10 $\pm$ 0.32} & \textbf{0.93 $\pm$ 0.07} & 0.0146 $\pm$ 0.0091 \\ \hline
        iris 
          & IMLI & \textbf{2.00 $\pm$ 0.00} & 7.60 $\pm$ 1.35 & 4.50 $\pm$ 1.08 & \textbf{0.90 $\pm$ 0.08} & \textbf{0.0051 $\pm$ 0.0010} \\
        ~ & IMLIB & \textbf{2.00 $\pm$ 0.00} & \textbf{4.90 $\pm$ 1.20} & \textbf{2.50 $\pm$ 0.71} & 0.84 $\pm$ 0.12 & 0.0523 $\pm$ 0.0378 \\ \hline
        parkinsons 
          & IMLI & \textbf{2.00 $\pm$ 0.00} & 5.00 $\pm$ 2.05 & 2.90 $\pm$ 1.37 & \textbf{0.80 $\pm$ 0.07} & \textbf{0.0223 $\pm$ 0.0033} \\
        ~ & IMLIB & \textbf{2.00 $\pm$ 0.00} & \textbf{3.00 $\pm$ 1.41} & \textbf{1.60 $\pm$ 0.84} & 0.79 $\pm$ 0.06 & 0.0631 $\pm$ 0.0263 \\ \hline
        ionosphere 
          & IMLI & \textbf{2.90 $\pm$ 0.32} & 12.00 $\pm$ 1.63 & 5.20 $\pm$ 0.63 & \textbf{0.81 $\pm$ 0.05} & \textbf{0.0781 $\pm$ 0.0096} \\
        ~ & IMLIB & 3.00 $\pm$ 0.00 & \textbf{7.70 $\pm$ 3.02} & \textbf{2.70 $\pm$ 1.16} & 0.79 $\pm$ 0.04 & 0.2797 $\pm$ 0.1087 \\ \hline
        wdbc 
          & IMLI & \textbf{2.90 $\pm$ 0.32} & 8.70 $\pm$ 2.50 & 3.70 $\pm$ 1.34 & \textbf{0.89 $\pm$ 0.03} & \textbf{0.0894 $\pm$ 0.0083} \\
        ~ & IMLIB & 3.00 $\pm$ 0.00 & \textbf{5.30 $\pm$ 2.36} & \textbf{1.80 $\pm$ 0.79} & 0.86 $\pm$ 0.06 & 0.2172 $\pm$ 0.0800 \\ \hline
        transfusion 
          & IMLI & \textbf{1.00 $\pm$ 0.00} & 3.10 $\pm$ 0.88 & 3.10 $\pm$ 0.88 & \textbf{0.72 $\pm$ 0.08} & \textbf{0.0291 $\pm$ 0.0026} \\
        ~ & IMLIB & \textbf{1.00 $\pm$ 0.00} & \textbf{2.00 $\pm$ 0.82} & \textbf{2.00 $\pm$ 0.82} & 0.68 $\pm$ 0.08 & 0.5287 $\pm$ 0.3849 \\ \hline
        pima 
          & IMLI & \textbf{1.00 $\pm$ 0.00} & 5.10 $\pm$ 0.74 & 5.10 $\pm$ 0.74 & 0.68 $\pm$ 0.09 & \textbf{0.0412 $\pm$ 0.0032} \\
        ~ & IMLIB & \textbf{1.00 $\pm$ 0.00} & \textbf{1.90 $\pm$ 1.10} & \textbf{1.90 $\pm$ 1.10} & \textbf{0.74 $\pm$ 0.04} & 0.6130 $\pm$ 0.5093 \\ \hline
        titanic 
          & IMLI & \textbf{1.00 $\pm$ 0.00} & 6.90 $\pm$ 1.91 & 6.90 $\pm$ 1.91 & 0.71 $\pm$ 0.07 & \textbf{0.0684 $\pm$ 0.0040} \\
        ~ & IMLIB & \textbf{1.00 $\pm$ 0.00} & \textbf{1.70 $\pm$ 0.67} & \textbf{1.70 $\pm$ 0.67} & \textbf{0.75 $\pm$ 0.06} & 1.9630 $\pm$ 3.2705 \\ \hline
        depressed 
          & IMLI & \textbf{1.80 $\pm$ 0.42} & 7.50 $\pm$ 2.64 & 5.30 $\pm$ 1.89 & 0.74 $\pm$ 0.08 & \textbf{0.2041 $\pm$ 0.0059} \\
        ~ & IMLIB & 2.00 $\pm$ 0.00 & \textbf{6.20 $\pm$ 3.36} & \textbf{3.30 $\pm$ 1.95} & \textbf{0.79 $\pm$ 0.04} & 0.5175 $\pm$ 0.2113 \\ \hline
        mushroom 
          & IMLI & \textbf{2.90 $\pm$ 0.32} & 16.30 $\pm$ 2.91 & 8.20 $\pm$ 2.20 & \textbf{0.99 $\pm$ 0.01} & \textbf{0.3600 $\pm$ 0.0340} \\
        ~ & IMLIB & 3.00 $\pm$ 0.00 & \textbf{12.30 $\pm$ 7.24} & \textbf{4.30 $\pm$ 2.54} & 0.97 $\pm$ 0.03 & 2.3136 $\pm$ 0.6294 \\ \hline
    \end{tabular}
\end{table}

\begin{table}[h!]
    \centering
    \scriptsize
    \caption{
        Comparison between IMLI and IMLIB in different databases with the 
        IMLIB configuration that obtained the best result in terms of 
        accuracy. The column Training time represents the training time in seconds.
    }\label{Table: All results IMLIB}
    \begin{tabular}{|c|c|c|c|c|c|c|}
    \hline 
        Databases & Models & \begin{tabular}{c} Number of\\ rules \end{tabular} & $|\textbf{R}|$ & \begin{tabular}{c} Largest rule\\ size \end{tabular} & Accuracy & Training time \\ [1.0ex]\hline\hline
        lung cancer 
          & IMLIB & \textbf{2.00 $\pm$ 0.00} & \textbf{2.20 $\pm$ 0.63} & \textbf{1.10 $\pm$ 0.32} & \textbf{0.93 $\pm$ 0.07} & 0.0146 $\pm$ 0.0091 \\
        ~ & IMLI & \textbf{2.00 $\pm$ 0.00} & 3.60 $\pm$ 0.84 & 2.20 $\pm$ 0.63 & \textbf{0.93 $\pm$ 0.07} & \textbf{0.0062 $\pm$ 0.0016} \\ \hline
        iris 
          & IMLIB & 2.90 $\pm$ 0.32 & \textbf{6.80 $\pm$ 1.48} & \textbf{2.50 $\pm$ 0.53} & \textbf{0.90 $\pm$ 0.07} & 0.0373 $\pm$ 0.0095 \\
        ~ & IMLI & \textbf{2.50 $\pm$ 0.53} & 9.10 $\pm$ 1.91 & 4.80 $\pm$ 0.92 & 0.86 $\pm$ 0.09 & \textbf{0.0062 $\pm$ 0.0011} \\ \hline
        parkinsons 
          & IMLIB & \textbf{3.00 $\pm$ 0.00} & \textbf{4.90 $\pm$ 1.66} & \textbf{1.70 $\pm$ 0.67} & \textbf{0.82 $\pm$ 0.07} & 0.0868 $\pm$ 0.0510 \\
        ~ & IMLI & \textbf{3.00 $\pm$ 0.00} & 8.40 $\pm$ 1.90 & 3.70 $\pm$ 1.06 & 0.79 $\pm$ 0.07 & \textbf{0.0295 $\pm$ 0.0064} \\ \hline
        ionosphere 
          & IMLIB & \textbf{2.00 $\pm$ 0.00} & \textbf{5.00 $\pm$ 1.70} & \textbf{2.50 $\pm$ 0.85} & \textbf{0.82 $\pm$ 0.06} & 0.2002 $\pm$ 0.0725 \\
        ~ & IMLI & \textbf{2.00 $\pm$ 0.00} & 7.90 $\pm$ 1.79 & 4.90 $\pm$ 1.45 & 0.80 $\pm$ 0.07 & \textbf{0.0531 $\pm$ 0.0106} \\ \hline
        wdbc 
          & IMLIB & \textbf{1.00 $\pm$ 0.00} & \textbf{1.20 $\pm$ 0.42} & \textbf{1.20 $\pm$ 0.42} & \textbf{0.89 $\pm$ 0.04} & 0.0532 $\pm$ 0.0159 \\
        ~ & IMLI & \textbf{1.00 $\pm$ 0.00} & 2.50 $\pm$ 0.71 & 2.50 $\pm$ 0.71 & 0.86 $\pm$ 0.09 & \textbf{0.0357 $\pm$ 0.0048} \\ \hline
        transfusion 
          & IMLIB & \textbf{1.00 $\pm$ 0.00} & \textbf{1.70 $\pm$ 0.67} & \textbf{1.70 $\pm$ 0.67} & \textbf{0.72 $\pm$ 0.03} & 0.2843 $\pm$ 0.1742 \\
        ~ & IMLI & \textbf{1.00 $\pm$ 0.00} & 3.10 $\pm$ 0.74 & 3.10 $\pm$ 0.74 & 0.71 $\pm$ 0.06 & \textbf{0.0273 $\pm$ 0.0032} \\ \hline
        pima 
          & IMLIB & \textbf{1.00 $\pm$ 0.00} & \textbf{1.90 $\pm$ 1.10} & \textbf{1.90 $\pm$ 1.10} & \textbf{0.74 $\pm$ 0.04} & 0.6130 $\pm$ 0.5093 \\
        ~ & IMLI & \textbf{1.00 $\pm$ 0.00} & 5.10 $\pm$ 0.74 & 5.10 $\pm$ 0.74 & 0.68 $\pm$ 0.09 & \textbf{0.0412 $\pm$ 0.0032} \\ \hline
        titanic 
          & IMLIB & \textbf{1.00 $\pm$ 0.00} & \textbf{1.40 $\pm$ 0.97} & \textbf{1.40 $\pm$ 0.97} & \textbf{0.76 $\pm$ 0.08} & 0.8523 $\pm$ 1.7754 \\
        ~ & IMLI & \textbf{1.00 $\pm$ 0.00} & 6.80 $\pm$ 1.87 & 6.80 $\pm$ 1.87 & 0.68 $\pm$ 0.12 & \textbf{0.0649 $\pm$ 0.0047} \\ \hline
        depressed 
          & IMLIB & 3.00 $\pm$ 0.00 & \textbf{13.30 $\pm$ 5.12} & \textbf{4.70 $\pm$ 1.89} & \textbf{0.80 $\pm$ 0.04} & 0.7263 $\pm$ 0.1692 \\
        ~ & IMLI & \textbf{2.90 $\pm$ 0.32} & 14.80 $\pm$ 2.25 & 6.70 $\pm$ 1.70 & 0.69 $\pm$ 0.08 & \textbf{0.2520 $\pm$ 0.0140} \\ \hline
        mushroom 
          & IMLIB & \textbf{1.00 $\pm$ 0.00} & \textbf{6.70 $\pm$ 0.95} & \textbf{6.70 $\pm$ 0.95} & \textbf{0.99 $\pm$ 0.00} & 1.2472 $\pm$ 0.2250 \\
        ~ & IMLI & \textbf{1.00 $\pm$ 0.00} & 8.90 $\pm$ 1.10 & 8.90 $\pm$ 1.10 & \textbf{0.99 $\pm$ 0.01} & \textbf{0.1214 $\pm$ 0.0218} \\ \hline
    \end{tabular}
\end{table}

In Table \ref{Table: All results IMLI}, when considering parameter configurations 
that favor IMLI, we can see that IMLIB stands out in the size of the generated set 
of rules and in the size of the largest rule in datasets. Furthermore, our method 
achieved equal or higher accuracy compared to IMLI in four out of ten datasets. In 
datasets where IMLI outperformed IMLIB in terms of accuracy, our method exhibited a 
modest average performance gap of only three percentage points. Besides, IMLI 
outperformed our method in terms of training time in all datasets.

In Table \ref{Table: All results IMLIB}, when we consider parameter configurations 
that favor our method, we can see that IMLIB continues to stand out in terms of the 
size of the generated set of rules and the size of the largest rule in all 
datasets. Moreover, our method achieved equal or higher accuracy than IMLI in all 
datasets. Again, IMLI consistently demonstrated better training time performance 
compared to IMLIB across all datasets.

As an illustrative example of interpretability, we present a comparison of the 
sizes of rules learned by both methods in the Mushroom dataset. Table 
\ref{Table: Mushroom rules size} shows the sizes of rules obtained in all ten 
realizations of the experiment. We can observe that IMLIB consistently maintains a 
smaller and more balanced set of rules across the different realizations. This is 
interesting because unbalanced rules can affect interpretability. See realization 
$6$, for instance. The largest rule learned by IMLI has a size of $10$, nearly 
double the size of the remaining rules. In contrast, IMLIB learned a set of rules 
where the size of the largest rule is $6$ and the others have similar sizes. Thus, 
interpreting three rules of size at most $6$ is easier than interpreting a rule of 
size $10$. Also as illustrative examples of interpretability, we can see some sets 
of rules learned by IMLIB in Table \ref{Table: Set of rules examples}.

\begin{table}[t]
    \centering
    \caption{
        Comparison of the size of the rules generated in the ten 
        realizations of the Mushroom base from Table \ref{Table: All 
        results IMLI}. The configuration used was $lp = 16$, $k = 3$ and 
        $\lambda = 10$. As the value of $l$ used in IMLIB varies across the 
        realizations, column $l$ will indicate which was the value used in 
        each realization. In the column Rules sizes, we show the size of 
        each rule in the following format: ($|R_1|$, $|R_2|$, $|R_3|$). We 
        have highlighted in bold the cases where the size of $|R_o|$ is the 
        same or smaller in our model compared to IMLI.
    }\label{Table: Mushroom rules size}
    \begin{tabular}{|c|c|c|c|c|}
        \hline
            Realizations & $l$ & Models & Rules sizes \\
        \hline
        \hline
            1 & -- & IMLI & (4, 6, 5) \\
            ~ & 1 & IMLIB & (\textbf{1}, \textbf{1}, \textbf{1}) \\
        \hline
            2 & -- & IMLI & (6, 11, \textbf{0}) \\
            ~ & 8 & IMLIB & (\textbf{3}, \textbf{6}, 6) \\
        \hline
            3 & -- & IMLI & (\textbf{3}, 8, \textbf{5}) \\
            ~ & 5 & IMLIB & (5, \textbf{5}, \textbf{5}) \\
        \hline
            4 & -- & IMLI & (6, 4, 4) \\
            ~ & 2 & IMLIB & (\textbf{2}, \textbf{2}, \textbf{2}) \\
        \hline
            5 & -- & IMLI & (3, 3, 4) \\
            ~ & 2 & IMLIB & (\textbf{2}, \textbf{2}, \textbf{2}) \\
        \hline
            6 & -- & IMLI & (\textbf{5}, 10, \textbf{6}) \\
            ~ & 6 & IMLIB & (6, \textbf{5}, \textbf{6}) \\
        \hline
            7 & -- & IMLI & (\textbf{5}, 9, \textbf{3}) \\
            ~ & 8 & IMLIB & (8, \textbf{8}, 8) \\
        \hline
            8 & -- & IMLI & (\textbf{4}, 10, \textbf{3}) \\
            ~ & 6 & IMLIB & (5, \textbf{5}, 6) \\
        \hline
            9 & -- & IMLI & (9, \textbf{5}, \textbf{4}) \\
            ~ & 6 & IMLIB & (\textbf{6}, 6, 6) \\
        \hline
            10 & -- & IMLI & (4, 9, 5) \\
            ~ & 1 & IMLIB & (\textbf{1}, \textbf{1}, \textbf{1}) \\
        \hline
    \end{tabular}
\end{table}

\begin{table}[h!]
    \centering
    \caption{
        Examples of set of rules generated by IMLIB in some 
        tested databases.
    }\label{Table: Set of rules examples}
    \begin{tabular}{|c|c|c|c|c|}
        \hline
            Databases & Sets of rules \\
        \hline
        \hline
            lung cancer & \begin{tabular}{c}
                (\textit{AreaQ} $\leq$ 5.0 and \textit{Alkhol} $>$ 3.0)
            \end{tabular} \\
        \hline
            iris & \begin{tabular}{c}
                (\textit{petal length} $\leq$ 1.6 and \textit{petal width} 
                $>$ 1.3) or \\ (\textit{sepal width} $\leq$ 3.0 and 
                \textit{petal length} $\leq$ 5.1)
            \end{tabular} \\
        \hline
            parkinsons & \begin{tabular}{c}
                 (\textit{Spread2} $>$ 0.18 and \textit{PPE} $>$ 0.19) or \\
                 (\textit{Shimmer:APQ3} $>$ 0.008 and 
            \textit{Spread2} $\leq$ 0.28)
            \end{tabular} \\
        \hline
            wdbc & \begin{tabular}{c}
                (\textit{Largest area} $>$ 1039.5) or \\ 
                (\textit{Area} $\leq$ 546.3 and \textit{Largest concave points} 
                $\leq$ 0.07)
            \end{tabular} \\
        \hline
            depressed & \begin{tabular}{c}
                (\textit{Age} $>$ 41.0 and \textit{Living expenses} $\leq$ 
                26692283.0) or \\ (\textit{Education level} $>$ 8.0 and 
                \textit{Other expenses} $\leq$ 20083274.0)
            \end{tabular} \\
        \hline
    \end{tabular}
\end{table}

\section{Conclusion}

In this work, we present a new incremental model for learning interpretable and 
balanced rules: IMLIB. Our method leverages the strengths of SQFSAT, which 
effectively constrains the size of rules, while incorporating techniques from IMLI, 
such as incrementability, cost for classification errors, and minimization of the 
set of rules. Our experiments demonstrate that the proposed approach generates 
smaller and more balanced rules than IMLI, while maintaining comparable or even 
superior accuracy in many cases. We argue that sets of small rules with 
approximately the same size seem more interpretable when compared to sets with a 
few large rules. As future work, we plan to develop a version of IMLIB that can 
classify sets of samples with more than two classes, enabling us to compare this 
approach with multiclass interpretable rules from the literature 
\cite{ignatiev2018sat,yu2020computing}.


\bibliographystyle{splncs04}
\bibliography{mybibliography}

\begin{thebibliography}{10}
\providecommand{\url}[1]{\texttt{#1}}
\providecommand{\urlprefix}{URL }
\providecommand{\doi}[1]{https://doi.org/#1}

\bibitem{biran2017explanation}
Biran, O., Cotton, C.: Explanation and justification in machine learning: A
  survey. In: IJCAI-17 workshop on explainable AI (XAI). vol.~8, pp. 8--13
  (2017)

\bibitem{carleo2019machine}
Carleo, G., Cirac, I., Cranmer, K., Daudet, L., Schuld, M., Tishby, N.,
  Vogt-Maranto, L., Zdeborov{\'a}, L.: Machine learning and the physical
  sciences. Reviews of Modern Physics  \textbf{91}(4),  045002 (2019)

\bibitem{Dua:2019}
Dua, D., Graff, C.: {UCI} machine learning repository (2017),
  \url{http://archive.ics.uci.edu/ml}

\bibitem{ghassemi2021false}
Ghassemi, M., Oakden-Rayner, L., Beam, A.L.: The false hope of current
  approaches to explainable artificial intelligence in health care. The Lancet
  Digital Health  \textbf{3}(11),  e745--e750 (2021)

\bibitem{ghosh2022efficient}
Ghosh, B., Malioutov, D., Meel, K.S.: Efficient learning of interpretable
  classification rules. Journal of Artificial Intelligence Research
  \textbf{74},  1823--1863 (2022)

\bibitem{ghosh2019imli}
Ghosh, B., Meel, K.S.: {IMLI}: An incremental framework for {MaxSAT}-based
  learning of interpretable classification rules. In: Proceedings of the 2019
  AAAI/ACM Conference on AI, Ethics, and Society. pp. 203--210 (2019)

\bibitem{gunning2019xai}
Gunning, D., Stefik, M., Choi, J., Miller, T., Stumpf, S., Yang, G.Z.:
  {XAI—Explainable} artificial intelligence. Science robotics
  \textbf{4}(37),  eaay7120 (2019)

\bibitem{huang2021power}
Huang, H.Y., Broughton, M., Mohseni, M., Babbush, R., Boixo, S., Neven, H.,
  McClean, J.R.: Power of data in quantum machine learning. Nature
  communications  \textbf{12}(1), ~2631 (2021)

\bibitem{ignatiev2021reasoning}
Ignatiev, A., Marques-Silva, J., Narodytska, N., Stuckey, P.J.: Reasoning-based
  learning of interpretable {ML} models. In: IJCAI. pp. 4458--4465 (2021)

\bibitem{ignatiev2019rc2}
Ignatiev, A., Morgado, A., Marques-Silva, J.: {RC2}: an efficient {MaxSAT}
  solver. Journal on Satisfiability, Boolean Modeling and Computation
  \textbf{11}(1),  53--64 (2019)

\bibitem{ignatiev2018sat}
Ignatiev, A., Pereira, F., Narodytska, N., Marques-Silva, J.: A {SAT-based}
  approach to learn explainable decision sets. In: Automated Reasoning: 9th
  International Joint Conference, IJCAR 2018, Held as Part of the Federated
  Logic Conference, FloC 2018, Oxford, UK, July 14-17, 2018, Proceedings 9. pp.
  627--645. Springer (2018)

\bibitem{janiesch2021machine}
Janiesch, C., Zschech, P., Heinrich, K.: Machine learning and deep learning.
  Electronic Markets  \textbf{31}(3),  685--695 (2021)

\bibitem{jimenez2020drug}
Jim{\'e}nez-Luna, J., Grisoni, F., Schneider, G.: Drug discovery with
  explainable artificial intelligence. Nature Machine Intelligence
  \textbf{2}(10),  573--584 (2020)

\bibitem{kwekha2021coronavirus}
Kwekha-Rashid, A.S., Abduljabbar, H.N., Alhayani, B.: Coronavirus disease
  (covid-19) cases analysis using machine-learning applications. Applied
  Nanoscience pp. 1--13 (2021)

\bibitem{lakkaraju2016interpretable}
Lakkaraju, H., Bach, S.H., Leskovec, J.: Interpretable decision sets: A joint
  framework for description and prediction. In: Proceedings of the 22nd ACM
  SIGKDD international conference on knowledge discovery and data mining. pp.
  1675--1684 (2016)

\bibitem{malioutov2018mlic}
Malioutov, D., Meel, K.S.: {MLIC}: A {MaxSAT}-based framework for learning
  interpretable classification rules. In: Principles and Practice of Constraint
  Programming: 24th International Conference, CP 2018, Lille, France, August
  27-31, 2018, Proceedings. pp. 312--327. Springer (2018)

\bibitem{mita2020libre}
Mita, G., Papotti, P., Filippone, M., Michiardi, P.: {LIBRE}: Learning
  interpretable boolean rule ensembles. In: AISTATS. pp. 245--255. PMLR (2020)

\bibitem{rocha2018synthesis}
Rocha, T.A., Martins, A.T.: Synthesis of quantifier-free first-order sentences
  from noisy samples of strings. In: 2019 8th Brazilian Conference on
  Intelligent Systems (BRACIS). pp. 12--17. IEEE (2019)

\bibitem{rocha2019synthesis}
Rocha, T.A., Martins, A.T., Ferreira, F.M.: Synthesis of a {DNF} formula from a
  sample of strings using {E}hrenfeucht--{F}ra{\"\i}ss{\'e} games. Theoretical
  Computer Science  \textbf{805},  109--126 (2020)

\bibitem{sharma2020machine}
Sharma, A., Jain, A., Gupta, P., Chowdary, V.: Machine learning applications
  for precision agriculture: A comprehensive review. IEEE Access  \textbf{9},
  4843--4873 (2020)

\bibitem{tjoa2020survey}
Tjoa, E., Guan, C.: A survey on explainable artificial intelligence ({XAI}):
  Toward medical {XAI}. IEEE transactions on neural networks and learning
  systems  \textbf{32}(11),  4793--4813 (2020)

\bibitem{vilone2021notions}
Vilone, G., Longo, L.: Notions of explainability and evaluation approaches for
  explainable artificial intelligence. Information Fusion  \textbf{76},
  89--106 (2021)

\bibitem{yan2020machine}
Yan, L., Zhang, H.T., Goncalves, J., Xiao, Y., Wang, M., Guo, Y., Sun, C.,
  Tang, X., Jing, L., Zhang, M., et~al.: An interpretable mortality prediction
  model for covid-19 patients. Nature machine intelligence  \textbf{2}(5),
  283--288 (2020)

\bibitem{yu2020computing}
Yu, J., Ignatiev, A., Stuckey, P.J., Le~Bodic, P.: Computing optimal decision
  sets with {SAT}. In: Principles and Practice of Constraint Programming: 26th
  International Conference, CP 2020, Louvain-la-Neuve, Belgium, September
  7--11, 2020, Proceedings 26. pp. 952--970. Springer (2020)

\end{thebibliography}

\end{document}